\documentclass[10pt,letterpaper,conference]{ieeeconf}
\usepackage[latin9]{inputenc}
\usepackage{amsmath}
\usepackage{graphicx}
\usepackage{hyperref}

\makeatletter

\pdfpageheight\paperheight
\pdfpagewidth\paperwidth

\usepackage{epstopdf}

\usepackage{multicol}

\makeatother

\begin{document}
\thispagestyle{empty}\pagestyle{empty}
\title{\textbf{Timely Negotiation and Correction of Shared Intentions With
Body Motion}}
\author{Raphael Deimel}
\thanks{All authors are with MTI-engAge Lab, TU Berlin}
\thanks{We gratefully acknowledge financial support for the project MTI-engAge
(16SV7109) by BMBF}
\maketitle
\begin{abstract}
Current robot architectures for modeling interaction behavior are not well suited to the dual task of sequencing discrete actions and incorporating information instantly. Additionally, for communication based on body motion, actions also serve as cues for negotiating interaction alternatives and to enable timely interventions. The paper presents a dynamical system based on the stable heteroclinic channel network, which provides a rich set of parameters to isntantly modulate motions, while maintaining a compact state graph abstraction suitable for reasoning, planning and inference. 
\end{abstract}

\section{INTRODUCTION}

Body language -- the use of body motion and pose for the purpose
of communication -- is a fast, intuitive and widely available modality
for negotiating shared intentions in physical human-robot interaction,
especially for collaborative tasks. Usually, task goals such as handing
over an object can be achieved in several ways But to succeed, both
parties have to agree upon a mutually consistent course of actions~\cite{strabala_towards_2013}.
Traditionally, robots determine that course of actions at a specific
point in time (decision points) \emph{prior }to executing actions,
and only reconsider choices \emph{after }completing the action, a
behavior that follows directly from the use of state machines (e.g.
hybrid automata, MDPs, grid worlds) to structure interaction patterns.
While discrete state machines provide huge advantages for learning,
reasoning and planning, they also discretize time which makes them
particularly unsuited for acting smoothly and timely on continuous
streams of perceptual information. They are also unable to perform
a \emph{speculative execution} of actions, i.e. to start an action
(e.g. reach out for handover) for the purpose of signaling an assumed
or preferred course of action to the interaction partner without committing
to its completion, so that the outcome can still be negotiated. It
buys the robot time to observe reactions to its motion and react accordingly,
e.g. by aborting an action or by blending to another, alternative
action. This way, robot and human can quickly negotiate courses of
actions, and if guessed correctly the first time (a probable scenario
due to cultural norms and individual preferences), then no extra time
is spent on the negotiation at all, making the interaction fluent
and swift. Modulation of body motion can also be used to effectively
negotiate roles in interactions. By displaying decisive motion, the
robot implicitly claims a leading role in an interaction, i.e. to
determine the location of a handover. Conversely, displaying hesitant
or ambiguous motions the human to take the lead and determine the
location.

Although the behaviors mentioned above could greatly improve intuitiveness
of human-robot interaction, implementation with discrete state machines
is cumbersome, difficult, and often requires giving up their prime
advantage: having a small state space. POMDPs are able to recreate
some form gradual behavior e.g. by using the expectation of states
for blending goals~\cite{nikolaidis_human-robot_2017,dragan_generating_2013},
but they do not provide a notion of reversibility required to implement
speculative execution and a notion of time (i.e. action phases) for
continuous synchronization. With hybrid automata~\cite{van_der_schaft_modeling_2000},
controllers can provide continuous behavior, but nevertheless hybrid
automata require decisions to be instant and irreversible. Also, any
perception-mediated modification of time-related behavior, i.e. phase
of a motion or relative importance of motions, has to bypass the hybrid
automaton and be implemented within controllers. As a consequence,
controllers are not reusable across tasks, state is fragmented across
the system and consistent modification of state (i.e. for conditioning
and learning) is difficult to achieve. 

To remove these shortcomings we propose a novel system architecture
to replace hybrid automata for robot behavior synthesis, one which
behaves like a discrete state machine but actually is a continuous
dynamical system. Additionally, it provides consistent activation
weights and phase values for mixing and blending controllers. 

The key conceptual difference to hybrid automata is that transitions
are extended over time, are non-exclusive (if they share a common
predecessor state), have a phase, and are revertible (i.e. are not
Markovian). The semantics of a discrete state machine can be recovered
by including transitions with their preceding state. So methods that
require markovian states -- most planning, probabilistic reasoning
and learning algorithms -- stay applicable.

Implementation as a dynamical system ensures that all information
paths are time-continuous and analytically differentiable, a property
that may especially be interesting for end-to-end learning approaches
that need gradients for each component. But it also ensures that perceptual
information can be integrated into the system state at any rate, any
time.

For human-robot interaction specifically, we will show how the proposed
system enables the robot to negotiate shared intentions on-the-fly
using body motion, convey preferences (e.g. the propensity to lead
or follow), and to synthesize timely and gradual feedback to cues
from human body motion; all without compromising the simplicity of
individual actions.

In the following sections we will first describe the implementation
and then demonstrate its capabilities in an object handover scenario.

\section{Implementation}

The proposed system builds upon the work on sable heteroclinic channel
(SHC) networks~\cite{horchler_designing_2015,rabinovich_transient_2008}.
SHC networks are dynamical systems that have saddle points which can
be arbitrarily connected with limit cycles (heteroclinic channels).
If the saddle points are interpreted as states, then SHC networks
can be understood to act like a state machine. and used as such~\cite{horchler_designing_2015}.
Fig.~\ref{fig:Illustration-3states} illustrates the attractor of
the simplest possible SHC network coposed of three saddle points.
In this paper, we additionally interpret the heteroclinic channels
as representing transitions between states, propose a method to algebraically
partition the state space into individual states and transitions as
well as compute a phase variable for each individual transition. Further,
the differential equation is modified to provide a \emph{greediness
}factor that modifies behavior during transitions.

.

.

.

The so-called phase-state machine combines a set of algebraic equations
with existing work on stable heteroclinic channel networks (SHC).
A SHC network is an attractor in a high-dimensional, continuous state
space with a number of saddle points, and stable channels connecting
these saddle points. The main feature of a SHC network is the straightforward
computability of the system matrices from a desired state transition
matrix, and that each saddle point is located along an exclusive coordinate
axis. It is important to realize another property of SHC networks
though: channels always lie in the plane spanned by the coordinate
axes of the preceding and succeeding saddle point, i.e. any transition
can be completely characterized by a rather simple projection into
a two-dimensional space. In a similar fashion, activation of a state
(represented by a saddle point) can be characterized by the distance
along a single dimension, due to the fact that the saddle point coordinates
form an orthonormal basis of the system state. This enables us to
compute from the state vector two properties: the activation of any
transition, and the phase of any transition. 

TODO: insert phase, activation computation

\subsection{Formal Definition}

Let $x$ be an $n$-dimensional vector that evolves according to this
differential equation\footnote{$\circ$ will be used throughout the paper to denote element-wise
multiplication (Hadamard product)}:

\begin{equation}
\dot{x}=x\circ\left(\alpha+\left(\rho_{o}+\rho_{\Delta}\circ\left(T+G\right)\right)\cdot x^{\gamma}\right)\cdot\eta(t)+\textrm{\ensuremath{\dot{\delta}}(t)+\ensuremath{\epsilon\cdot}\ensuremath{\mathcal{W}}(t)}\label{eq:DE_core_derivative}
\end{equation}
Compared to the equation used in \cite{horchler_designing_2015},
we added the exponent $\gamma$, explicitly introduce the state transition
matrix $T$ ($T_{ji}=1$if transition $i\rightarrow j$ exists, 0
otherwise), added a ``greediness'' matrix $G$, and added a scalar
$\eta(t)$ to adjust the speed at which $x$ evolves. The parameters
$\alpha$, $\rho_{0}$and $\rho_{\Delta}$ are chosen such that $n$
saddle points occur, each one placed on its exclusive coordinate axis.
The signal $\textrm{\ensuremath{\dot{\delta}}(t)}$ is used to selectively
push the system away from saddle points and $\textrm{\ensuremath{\epsilon\cdot}\ensuremath{\mathcal{W}}(t)}$
adds stochastic noise with zero mean.

\paragraph{Matrices $\rho_{o}$and $\rho_{\Delta}$ }

The $n\times n$ matrices $\rho_{0}$ and $\rho_{\Delta}$ are constructed
from three parameter vectors~\cite{horchler_designing_2015}: $\alpha$
(growth rates), $\beta$ (saddle point positions), and $\nu$ (saddle
point shapes):

\[
\rho_{o}=\left[\alpha\otimes\beta^{-1}\right]\circ\left[I-1-\alpha\otimes\alpha^{-1}\right]
\]
\[
\rho_{\Delta}=\left(\alpha\circ\left(1+\nu^{-1}\right)\right)\otimes\beta^{-1}
\]

The matrices are chosen such that the matrix $\rho$ constructed by
Eq.~5 in \cite{horchler_designing_2015} can be computed as $\rho=-\rho_{o}-T\circ\rho_{\Delta}$.
The advantage of the given formulation is that $\rho_{0}$ and $\rho_{\Delta}$
do not change when transition matrix $T$or greediness matrix $G$
is modified. For convenience, we can fix many parameters to obtain
a canonical system:
\begin{align*}
\alpha_{i} & =\alpha_{0} &  & \textrm{(growth rates)}\\
\beta_{i} & =1.0 &  & \textrm{(position of saddle point)}\\
\nu_{i} & =1.0 &  & \textrm{(channel asymmetry)}
\end{align*}

\begin{figure}
\includegraphics[width=1\columnwidth]{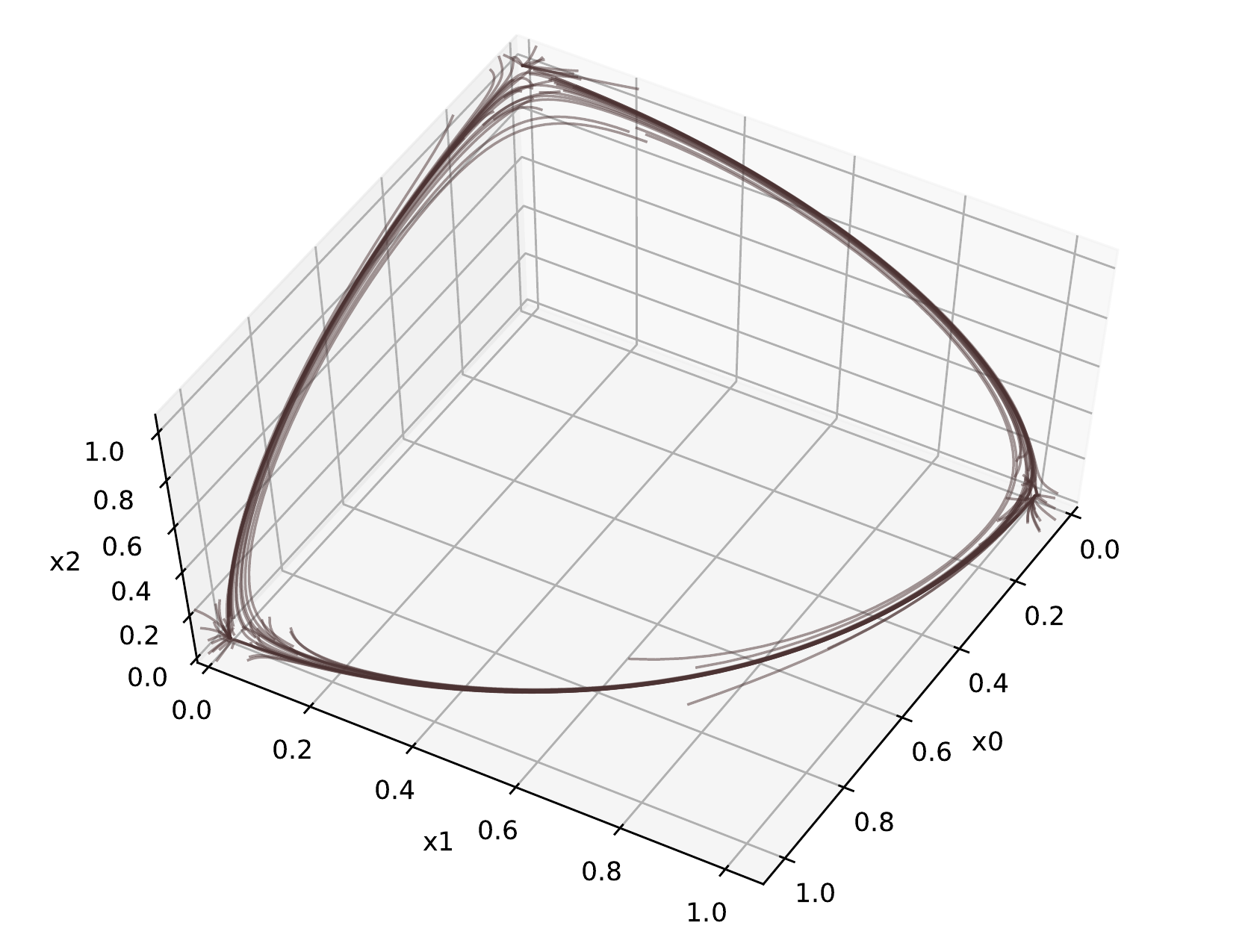}

\caption{Illustration of the vector field of a canonical system with three
dimensions, three saddle points and three heteroclinic channels. Saddle
points located at each coordinate axis, connected into a cycle $1\rightarrow2\rightarrow3\rightarrow1$.\label{fig:Illustration-3states}}
\end{figure}

To illustrate, the matrices for the system in Fig.~\ref{fig:Illustration-3states}
are:

\[
\rho_{0}=\left[\begin{array}{ccc}
-\alpha_{0} & -2\alpha_{0} & -2\alpha_{0}\\
-2\alpha_{0} & -\alpha_{0} & -2\alpha_{0}\\
-2\alpha_{0} & -2\alpha_{0} & -\alpha_{0}
\end{array}\right]
\]
\[
\rho_{\Delta}=\left[\begin{array}{ccc}
2\alpha_{0} & 2\alpha_{0} & 2\alpha_{0}\\
2\alpha_{0} & 2\alpha_{0} & 2\alpha_{0}\\
2\alpha_{0} & 2\alpha_{0} & 2\alpha_{0}
\end{array}\right]
\]
\[
T=\left[\begin{array}{ccc}
0 & 0 & 1\\
1 & 0 & 0\\
0 & 1 & 0
\end{array}\right]
\]

\paragraph*{Channel location}

The factor $\gamma$ determines the distance of the attractor to the
vector space origin. With $\gamma=1$ channels approximately maintain
constant $L^{1}$ distance (as used in \cite{horchler_designing_2015}),
whereas with $\gamma=2$ they approximately maintain constant $L^{2}$
distance (assuming a canonical system). The latter causes the attractor
to lie on a hypersphere. For the canonical system, we chose $\gamma=2$.

\subsection{Activations and Phases}

The SHC network provides the notion of states (saddle points) and
transitions (stable heteroclinic channels). In order to algebraically
partition the vector space of $x$ into regions for each state and
each possible transition, we can leverage two mathematical properties
of the system.

First, the coordinate vectors of each state/saddle point form an orthonormal
basis. From this follows that the channels are located on the plane
spanned by the basis vectors of predecessor and successor state, as
can also be seen in Fig.~\ref{fig:Illustration-3states}.

Second, the coordinate vector of each state is sparse, all but one
coordinates are zero. From this follows that functions specific to
a state or transition can be computed from specific elements of $x$.

\paragraph{Activation values of states and transitions}

From these insights we can devise an ``activation'' value for each
transition $i\rightarrow j$, based on its respective successor and
predecessor coordinate values and the norms of $x$:

\begin{eqnarray}
\Lambda^{\textrm{transitions}} & = & \frac{16\cdot x\otimes x\cdot\vert x^{2}\vert}{\left(x\otimes\boldsymbol{1}+\boldsymbol{1}\otimes x\right)^{4}+\vert x\vert^{4}}\circ T\label{eq:lambda_transition}
\end{eqnarray}

\begin{figure}
\includegraphics[width=1\columnwidth]{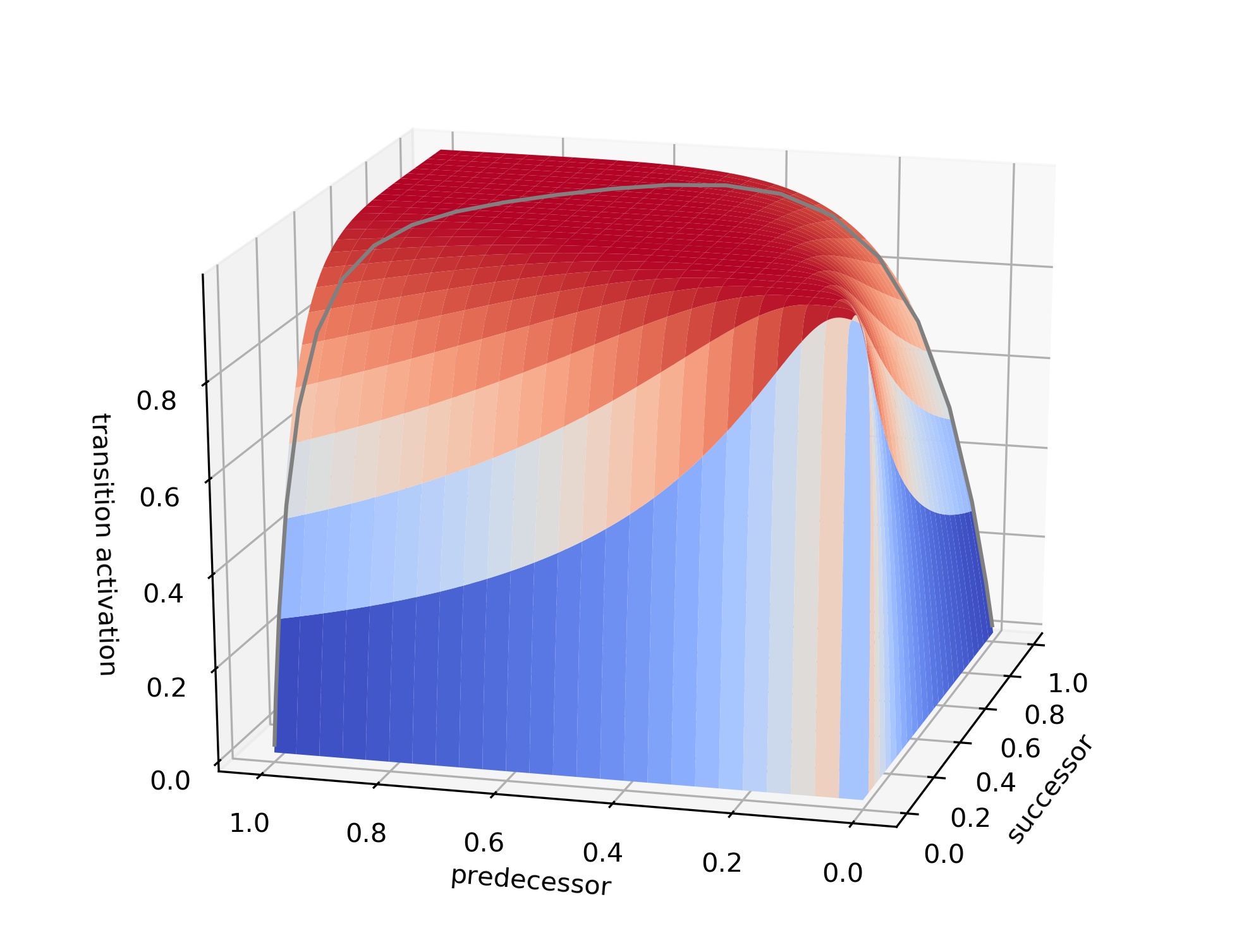}

\caption{Illustration of activation $\Lambda_{ji}$ in the plane of the related
transition $i\rightarrow j$. Grey line indicates the location of
heteroclinic channel for $\gamma=2$.}
\label{activation_faction_illustration}
\end{figure}

The function\footnote{$\otimes$ denotes the outer vector product.}
is chosen such that elements are limited to the range of $[0.0\ldots1.0]$,
and invariant to scaling $x$. Fig.\ref{activation_faction_illustration}
illustrates the function value for a single active transition $i\rightarrow j$
w.r.t. coordinates $x_{i}$ and $x_{j}$. $\Lambda^{\textrm{transitions}}$
is sparse in the sense that only few transitions are active at any
time. If more than one transition is active, then $\sum\Lambda^{\textrm{transitions}}\approx1.0$
(for systems with $\gamma=2$). Because of this, $\Lambda^{\textrm{transitions}}$
can also be understood as a weight matrix.

For the states, activation is computed from the residual of the transition
activations, so that all activation values sum up to $1.0$. Additionally,
$x$ is squared to ensure sparseness of the state activation values
and hence mutual exclusiveness:

\begin{eqnarray}
\lambda^{\textrm{states}} & = & x^{2}\cdot(1-\frac{\sum\Lambda^{\textrm{transitions}}}{\vert x^{2}\vert})\label{eq:lambda_states}
\end{eqnarray}

And as the diagonal of $\Lambda^{\textrm{transitions}}$ is semantically
not meaningful, we can combine all transition and state activations
into a single activation matrix $\Lambda$:
\[
\Lambda_{ji}=\begin{cases}
\lambda_{ji}^{\textrm{transitions}} & j\neq i\\
\lambda_{i}^{\textrm{states}} & j=i
\end{cases}
\]

\begin{figure}
\includegraphics[width=1\columnwidth]{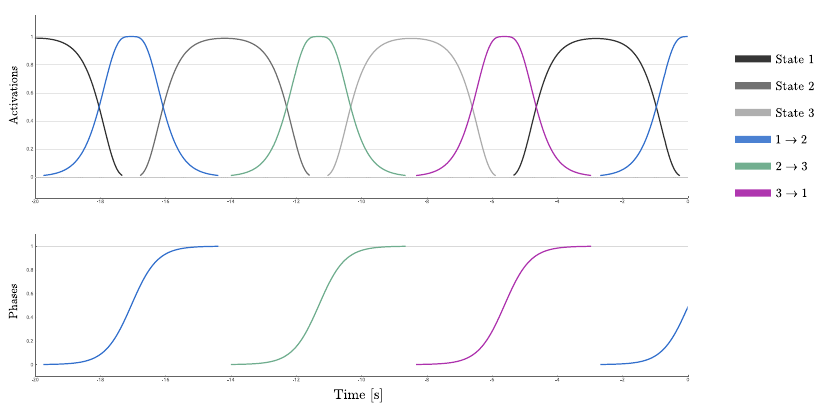}

\caption{Activations and phases resulting from the 3-state system shown in
Fig.~\ref{fig:Illustration-3states}. For clarity, values are not
drawn when the related activation is less than 0.01.\label{fig:Activations-and-phases-3states}}
\end{figure}

Fig.~\ref{fig:Activations-and-phases-3states} shows an example of
the resulting set of activation values for the minimal three-state
system illustrated in Fig.~\ref{fig:Illustration-3states} (using
$\alpha_{0}=10$, $\dot{\delta}=5\cdot10^{-5}$) .

\begin{figure}
\includegraphics[width=1\columnwidth]{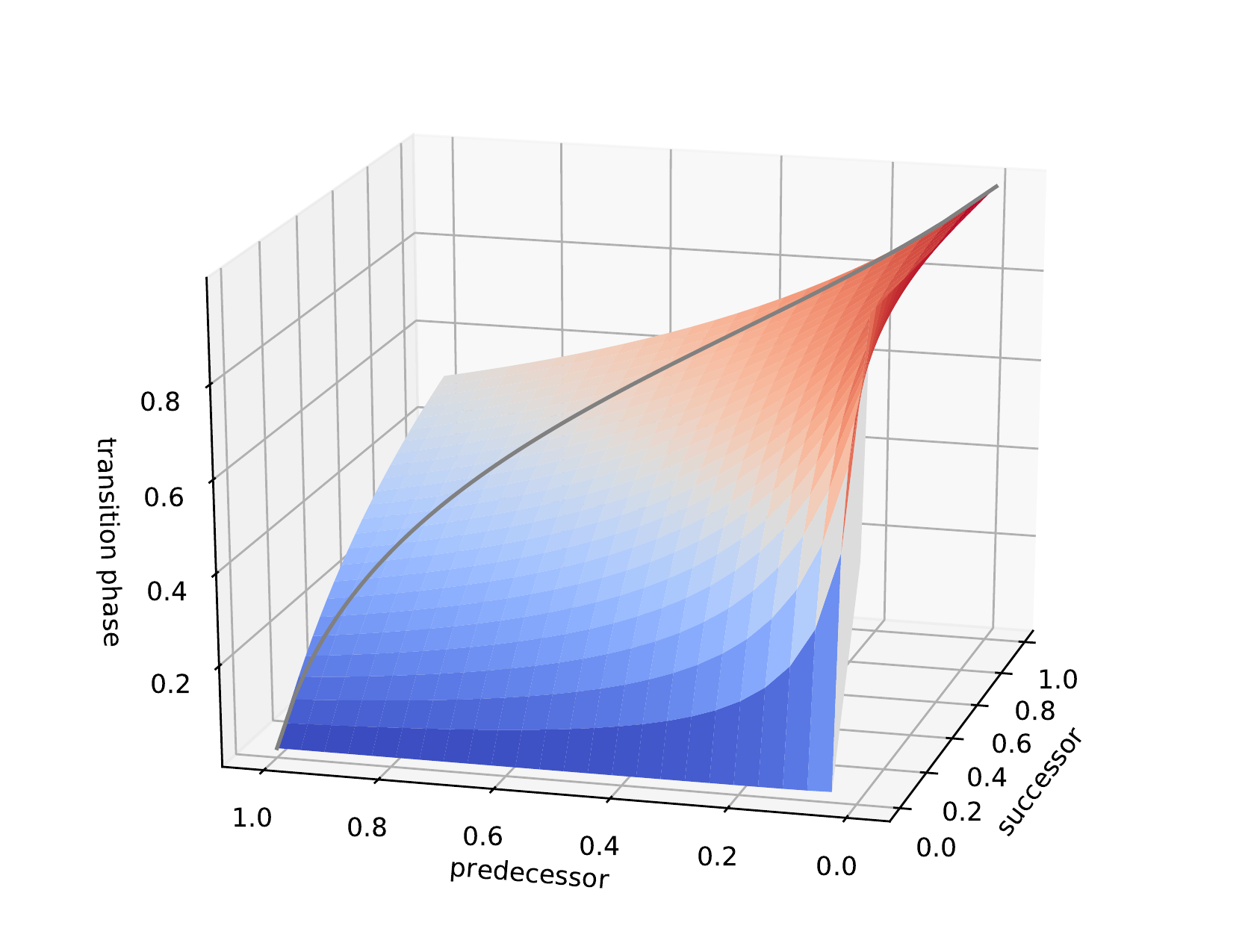}

\caption{Illustration of phase $\Psi_{ji}$ in the plane of the related transition
$i\rightarrow j$. Grey line indicates the location of heteroclinic
channel for $\gamma=2$.}
\label{phase_faction_illustration}
\end{figure}

\paragraph{Transition Phases}

Different to (markovian) states, transitions have a notion of time
and progress, i.e. they posses a phase. As channels are located on
a two-dimensional plane spanned by two coordinate axes, we can compute
a phase for each possible transition $i\rightarrow j$:

\begin{eqnarray}
\Phi_{ji} & = & \frac{\vert x_{j}\vert}{\vert x_{i}\vert+\vert x_{j}\vert}\label{eq:phaseprogressionmatrix_simple}
\end{eqnarray}

The shape of the function is illustrated in~\ref{activation_faction_illustration},
and yields values in the range $\left[0\ldots1\right]$. Note that
$\Phi_{ji}$ is only meaningful when transition $i\rightarrow j$
is active, i.e. when $\vert x_{i}\vert+\vert x_{j}\vert\gg0$. Fig.~\ref{fig:Activations-and-phases-3states}
illustrates the phase over time.

\subsubsection{Composition of Motion}

So far, we established a dynamical system that provides us with a
consistent set of activation values for transitions and states, and
with phases for transitions. Eqs.~\ref{eq:lambda_transition} and
\ref{eq:lambda_states} are chosen such that $\sum\Lambda=1$, therefore
$\Lambda$ can be directly used for weighted averaging of control
goals associated with each state and each transition. In terms of
control, states and transitions have to be treated differently though.
States are phase-less, so we can only associate static control goals
with them. Transitions, on the other hand,  have a phase, so we can
also associate phase-parameterized movement primitives with them,
such as DMPs~\cite{paraschos_probabilistic_2013,schaal_dynamic_2006}
and ProMPs~\cite{paraschos_probabilistic_2013}, or simply planned
trajectories. For the full system demonstration, we use the ProMP
framework to learn and reproduce movements during transitions. In
order to enable composition using the mixing method for ProMPs~\cite{paraschos_probabilistic_2013},
state goals are defined as static normal \emph{distribution} over
position and velocity. It is important to note that even though usually
only one or two control goals are activated, multiple goals may be
active, e.g. when competing transitions (with common predecessor state)
become active, or when subsequent transitions are blended into each
other because of large values in $\dot{\delta}$.

\subsection{Inputs to influence system behavior}

Terms of Eq.~\ref{eq:DE_core_derivative} is chosen such that some
of them can be used as inputs to effect certain behaviors. The transition
matrix T is used to define which transitions exist, and can be updated
during execution of the system, if desired. Matrix G is used to adjust
the behavior for active, competing transitions and for pausing or
aborting transitions. Vector $\dot{\delta}$ determines, when a state
is left and which transition(s) is activated. The factor $\eta$ speeds
up or slows down the system dynamics, which can be used for e.g. synchronization
by entraining.

\paragraph{Causing transitions}

When the system is exactly on a saddle point, e.g. $x=(1,0,0)$, then
the system can potentially stay in this state forever. In order to
cause a transition, a small positive velocity bias $\dot{\delta_{j}}$
can be added, which pushes the system towards successor state $j$,
or a negative $\dot{\delta_{j}}$ to avoid it. Sometimes though, this
level of granularity is not enough, and we want to set the velocity
bias for each transition specifically. We can define an input biases
matrix $B$ where each element $B_{ji}$ corresponds to the bias towards
state $j$ in state $i$. A resolved vector $\dot{\delta}$ can then
be computed with $\Lambda$ :

\begin{equation}
\dot{\delta}=\left(\Lambda\circ B\right)\cdot x\label{eq:kd-1}
\end{equation}

The matrix B elements are the equivalent of control switch conditions
in hybrid automata, i.e. B can be used to synchronize on events and
to select one out of several successor states. But it also can be
used to implement timeout conditions by using small values whose effect
gradually accumulates. Indeed, B was set to a small positive value
for generating for generating timeouts to the states in Figs.~\ref{fig:Illustration-3states}
and\ref{fig:Activations-and-phases-3states}. If needed, bias values
for specific durations can be estimated analytically~\cite{horchler_designing_2015}.

Another option to cause transitions is to add stochastic velocity
noise via $\epsilon$. In contrast to $\dot{\delta}$ it will cause
the system to transition after a random amount of time. This might
be useful in some interaction scenarios (e.g. avoiding synchronous
access to a resource, or exploratory behavior). Usually though, $\epsilon=0$.

\paragraph*{Transition Velocity}

A key advantage of the proposed system to hybrid automata is the ability
to continuously adjust the speed of a movement. In prior work, velocity
was adjusted by modifying the growth rate $\alpha$ \cite{horchler_designing_2015}.
Unfortunately though, stability considerations limit the range of
values that can be assigned to each $\alpha_{j}$. By using the activation
matrix $\Lambda$ though, we can modify the growth rate (and thus
speed of evolution) for each region in vector space independently:

\begin{equation}
\eta=2^{\sum\Lambda\circ A}\label{eq:transitionvelocityexponentsmatrix}
\end{equation}

Matrix $A$ contains factors for speeding up or slowing down each
transition and state relative to the ``default'' speed defined by
$\alpha_{o}$. The proposed approach, works well across several orders
of magnitude as it does not warp the saddle points. Unmodified system
behavior is obtained by setting $A=0$.

\begin{figure}
\includegraphics[width=1\columnwidth]{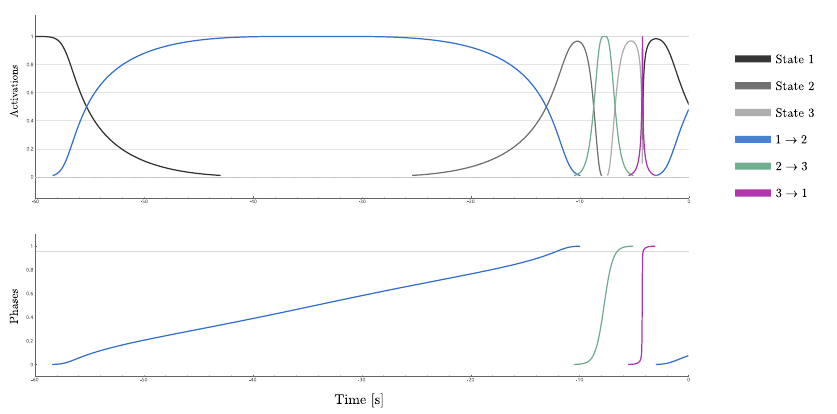}

\caption{Example of transition velocities ranging 3 orders of magnitude. $A_{21}=-5$,
$A_{32}=0$,$A_{13}=5$}
\end{figure}

\paragraph{Decisiveness and Hesitation}

\begin{figure}
\includegraphics[width=0.49\columnwidth]{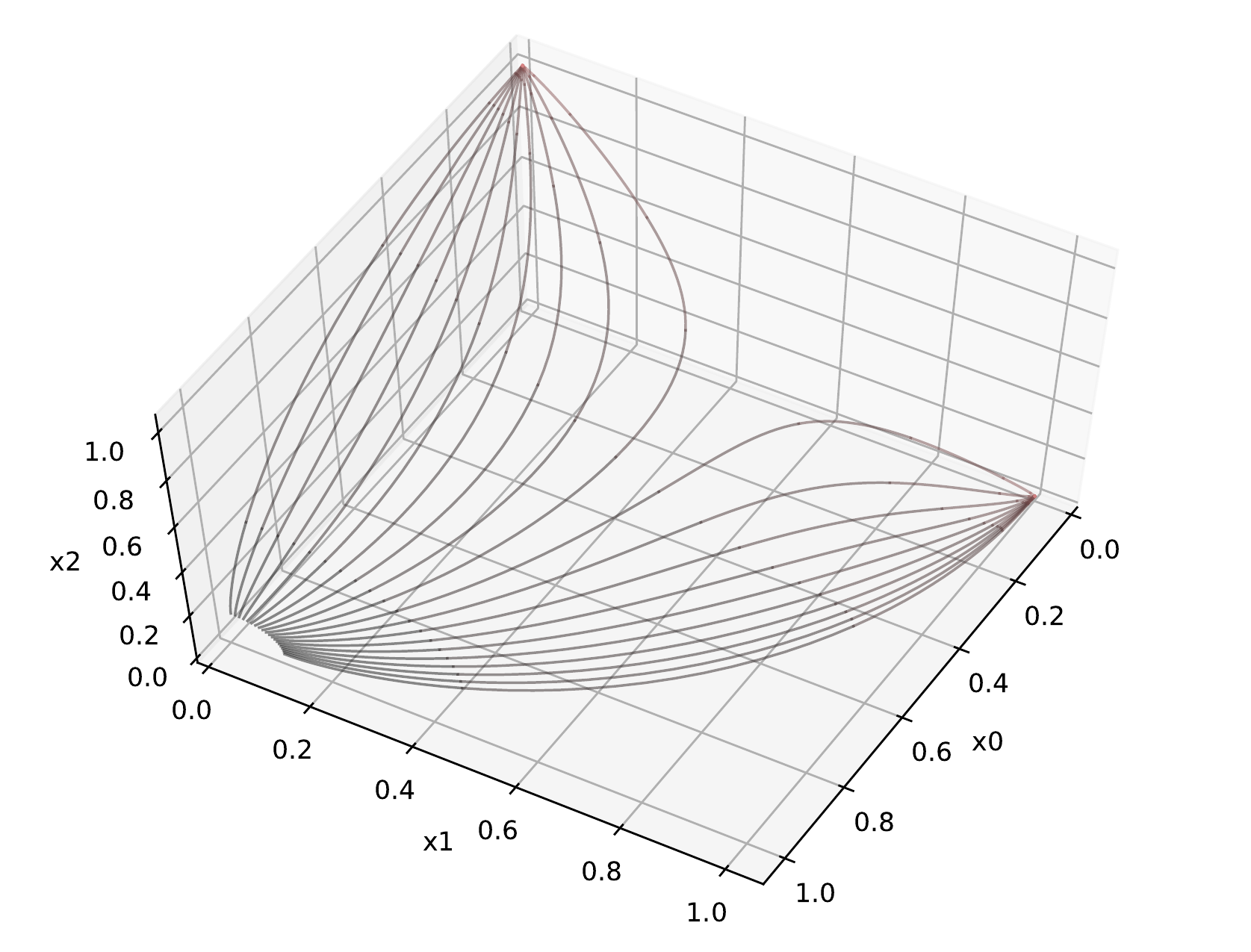}\includegraphics[width=0.49\columnwidth]{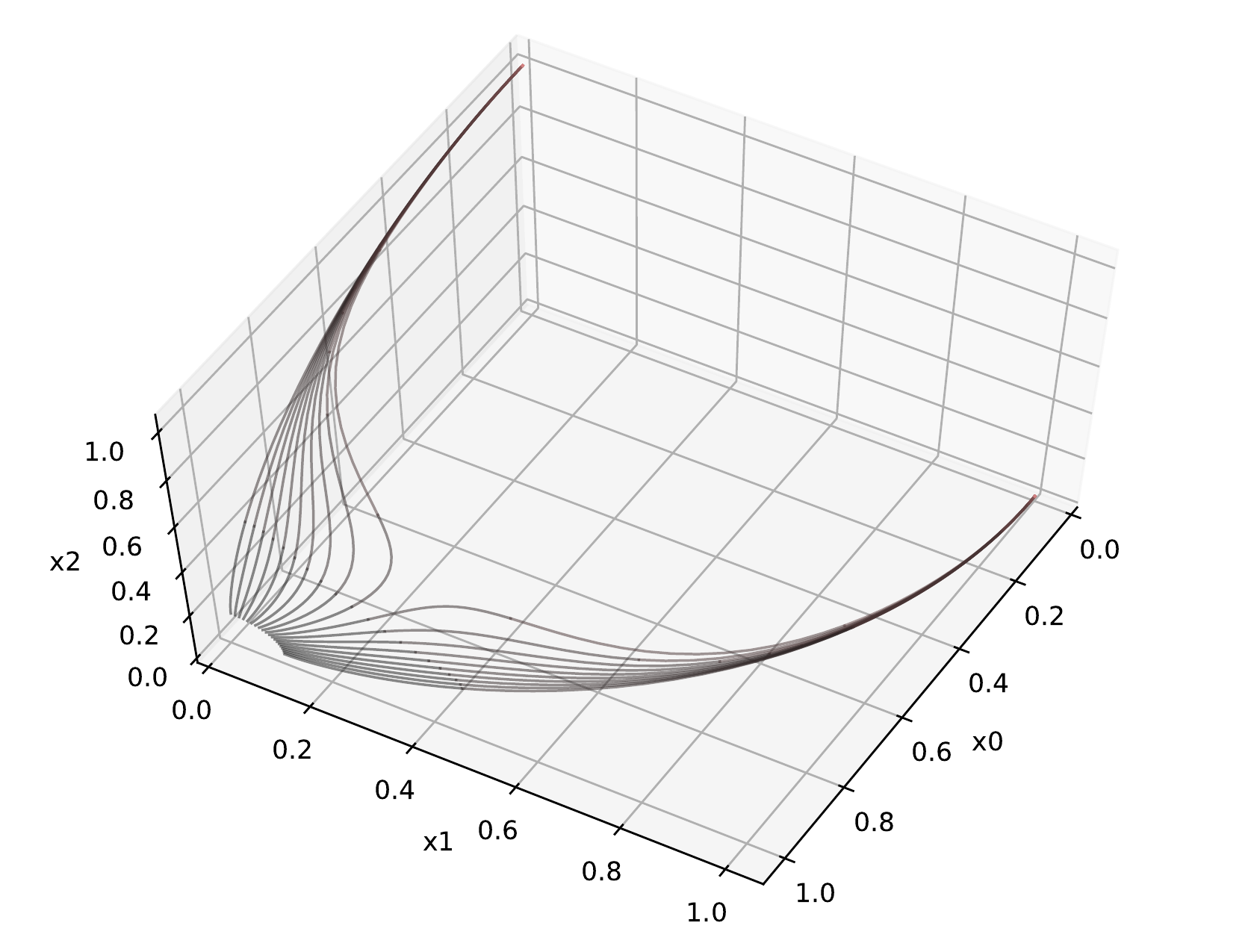}

\caption{Effect of changing the greediness parameter uniformly on two competing
transitions $0\rightarrow1$ and $0\rightarrow2$. (a) The system
can be reluctant to choose (original SHC behavior, $g=[1,1,1]$).
(b) It can be very decisive ($g=[8,8,8]$). (c) With $g=[0,0,0]$
ongoing transitions are halted. (d) With negative values transitions
are aborted and the the system returns to the predecessor state 0
($g=[-2,-2,-2]$). \label{fig:symmetric_greediness}}
\end{figure}

A unique feature of the proposed system is the ability to transition
from a predecessor state into the direction of several successor states
at once, by setting positive biases for transitions with common predecessor.
The attractor shape forces a decision at some point though and only
one transition completes, i.e. the system converges to one heteroclinic
channel, a behavior which ensures the mutual exclusivity of states.
The dynamic behavior of two competing heteroclinic channels is illustrated
in Fig.~\ref{fig:symmetric_greediness}a. Depending on $\dot{\delta}$,
the system state $x$ will first progress in a specific direction
on the hypersphere, but then trajectories will converge towards either
of the succeeding saddle points.

It turns out that this ``greediness'' of successor states can be
consistently modified by introducing a matrix $G$ in Eq.~\ref{eq:DE_core_derivative}:

\[
G=\left[T\circ\overrightarrow{G}-\overrightarrow{G}^{T}\circ T^{T}\right]-\left[TT^{T}\circ(1-I)\right]\circ\overleftrightarrow{G}
\]

The matrix $\overleftrightarrow{G}$ encodes competitive greediness,
i.e. mutual inhibition between competing successor states, while $\overrightarrow{G}$
encodes greediness w.r.t. the preceding state. When $\overleftrightarrow{G}=0$
and $\overrightarrow{G}=0$ the system behaves as in \cite{horchler_designing_2015}
and Fig.~\ref{fig:symmetric_greediness}a, when $\overrightarrow{G_{ji}}=-1$,
the gradient for the channel $i\rightarrow j$ is compensated, i.e.
the transition halts. If $\overrightarrow{G}=\overleftrightarrow{G}$,
then For simplicity we can define a single greediness vector $g$
with values for each (successor) state, from which we can construct
both matrices:
\[
\overrightarrow{G}_{ji}=\begin{cases}
-0.5 & g_{j}<-1\\
\frac{g_{j}-1}{2} & else\\
0 & g_{j}>1
\end{cases}
\]
and 
\[
\overleftrightarrow{G}_{ji}=1.5\cdot\frac{g_{j}-1}{2}-0.5\cdot\frac{g_{i}-1}{2}
\]

\begin{figure*}
\includegraphics[width=0.49\columnwidth]{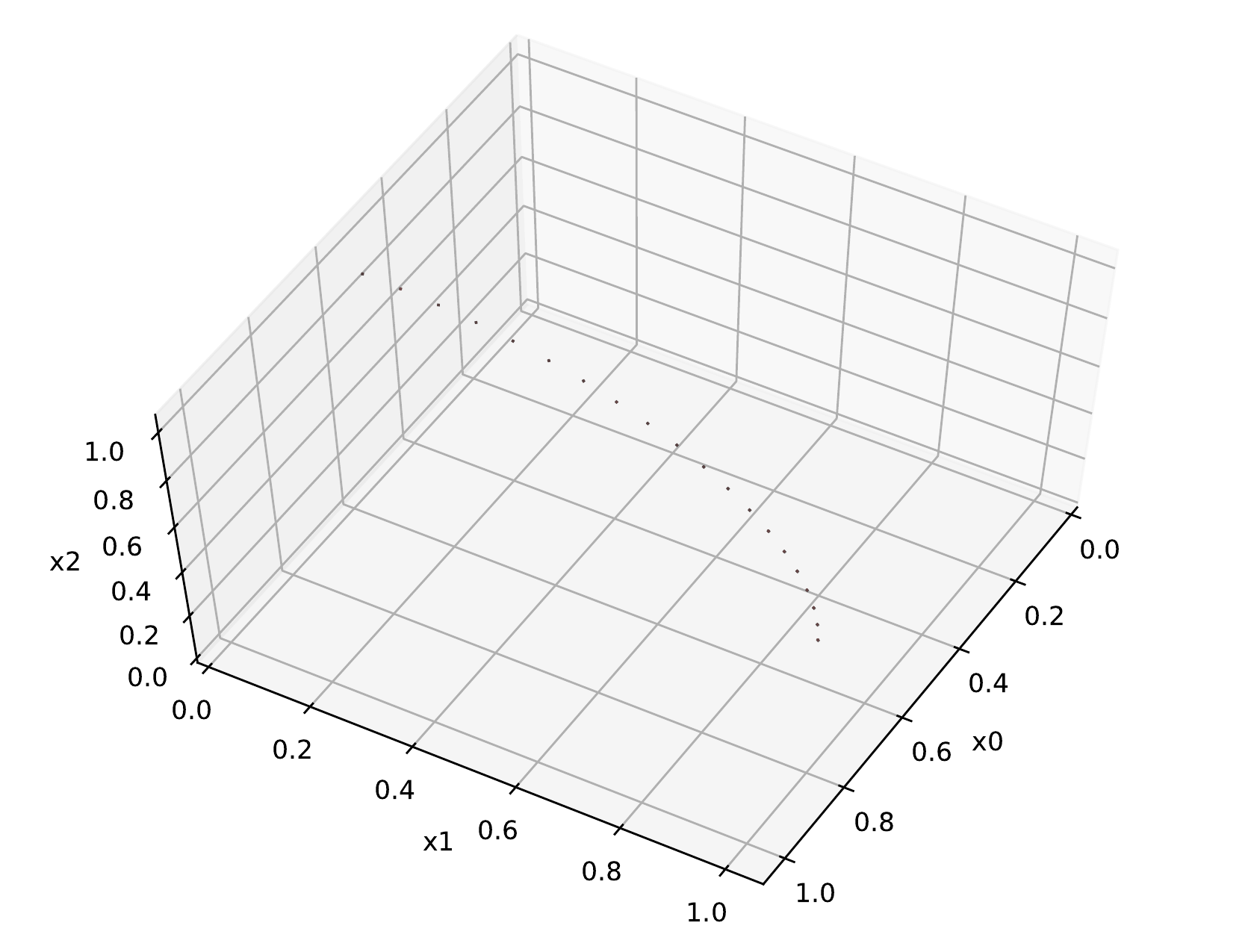}\includegraphics[width=0.49\columnwidth]{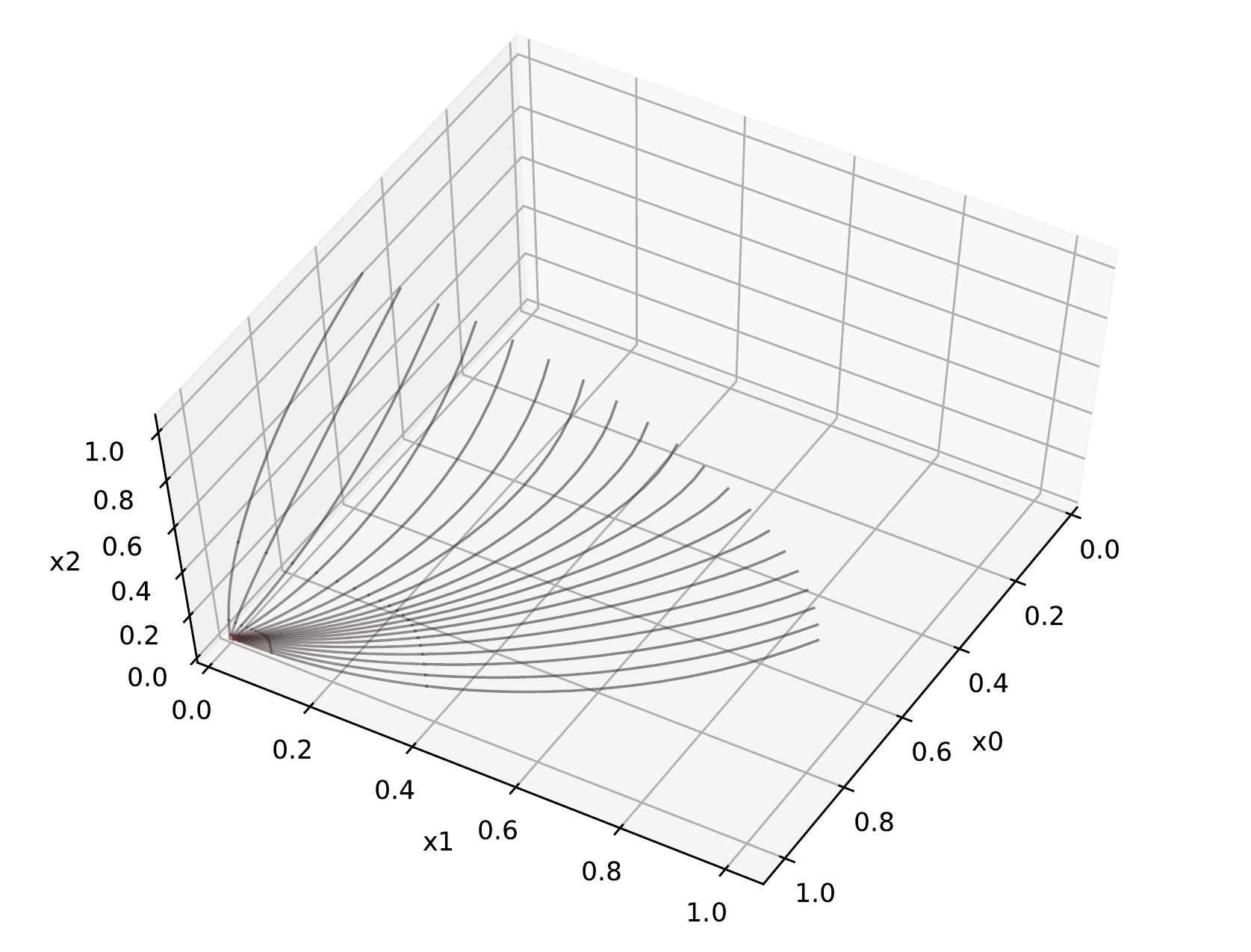}\includegraphics[width=0.49\columnwidth]{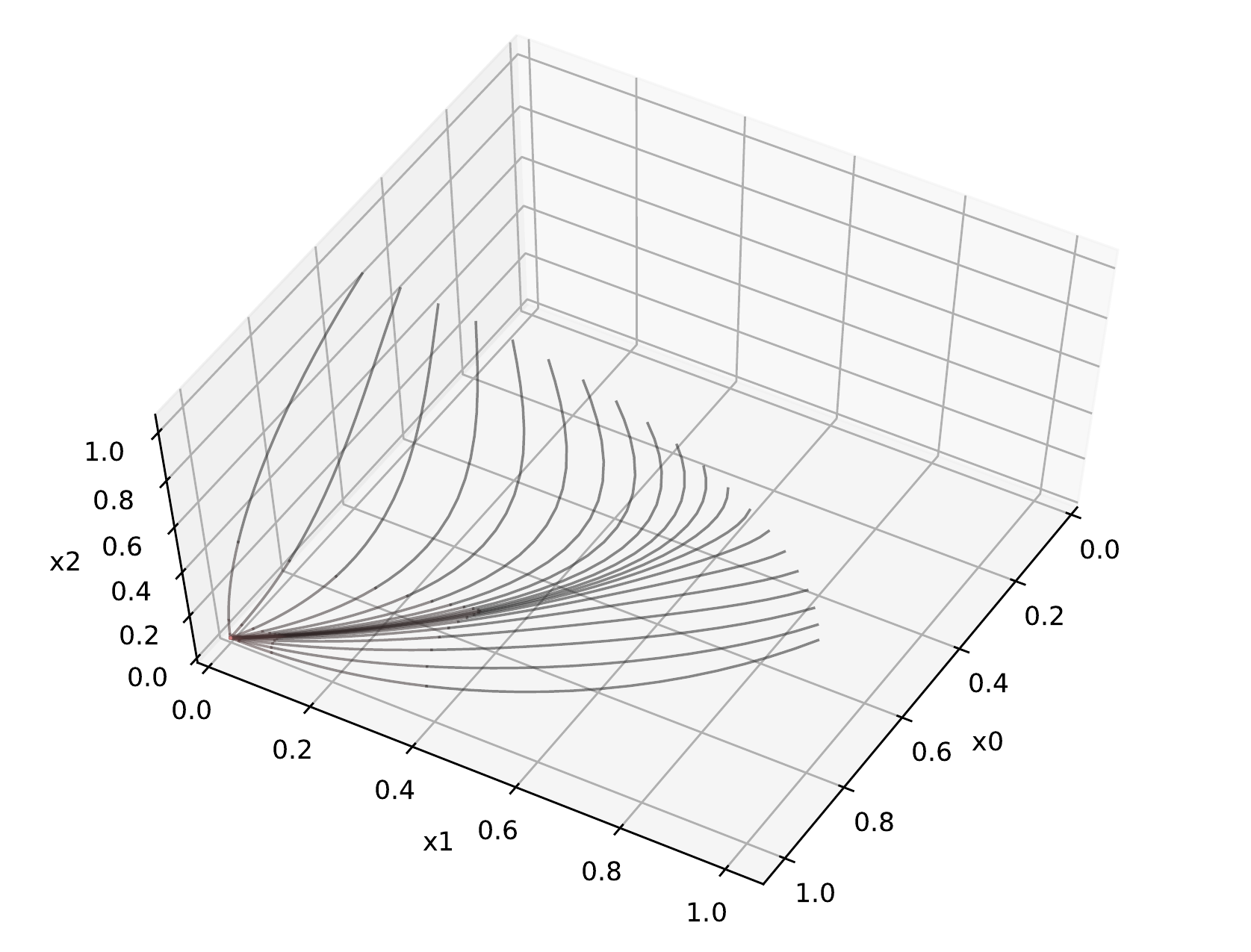}

\caption{Effects of non-positive greediness. (a) $g=[1,0,0]$ halts ongoing
transitions. (b) $g=[1,-1,-1]$ reverses ongoing transitions (c) $g=[1,-2,-2]$
reverses ongoing transitions and additionally balances them.\label{fig:symmetric_greediness-lessthan1}}
\end{figure*}

The equations are chosen such that the behavior of the original SHC
network is retrieved with $g_{j}=\text{1}$. (``default'' greediness).
With $g_{j}=0$ (Fig.~\ref{fig:symmetric_greediness-lessthan1}a),
the system will completely halt ongoing transitions towards state
$j$, i.e. the gradient along the heteroclinic channel drops to zero.
With negative $g_{j}$ (Fig.~\ref{fig:symmetric_greediness-lessthan1}b),
the gradient along the heteroclinic channel reverses, which moves
the system back to the preceding state.

The speed of transitions are not increased beyond the default speed
because $\overrightarrow{G}$ is clamped. Values beyond $\vert g_{j}\vert>1$
therefore only increases the competition between successor states.
The net effect is, that for large values of $g$ , the system becomes
very decision-happy (Fig.~\ref{fig:symmetric_greediness}b vs. Fig.~\ref{fig:symmetric_greediness}b)
and tries to converge towards a single transition early, while for
low values of $g$, the system is reluctant to decide. This effect
can be used to modulate the ambiguity of movements. If two expressive
movements are associated with two competing transitions, then large
values of $g$ will cause the system to avoid mixing movements, which
maintains their expressiveness. If $g$ is small then movements are
mixed according to the accumulated $\dot{\delta}$, creating an ambiguous
motion.

\paragraph*{Reconsidering Decisions}

\begin{figure*}
\includegraphics[width=0.33\textwidth]{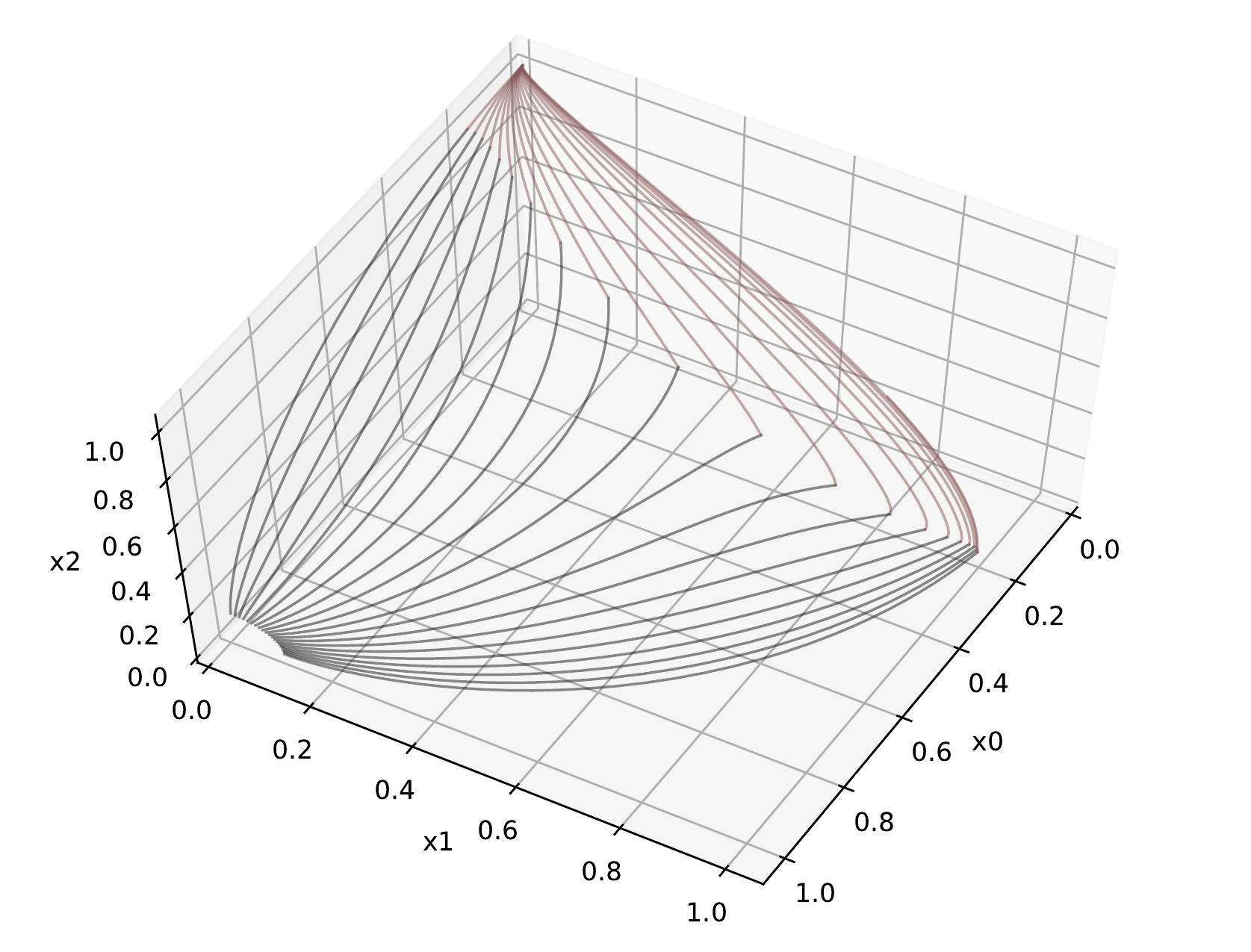}\includegraphics[width=0.33\textwidth]{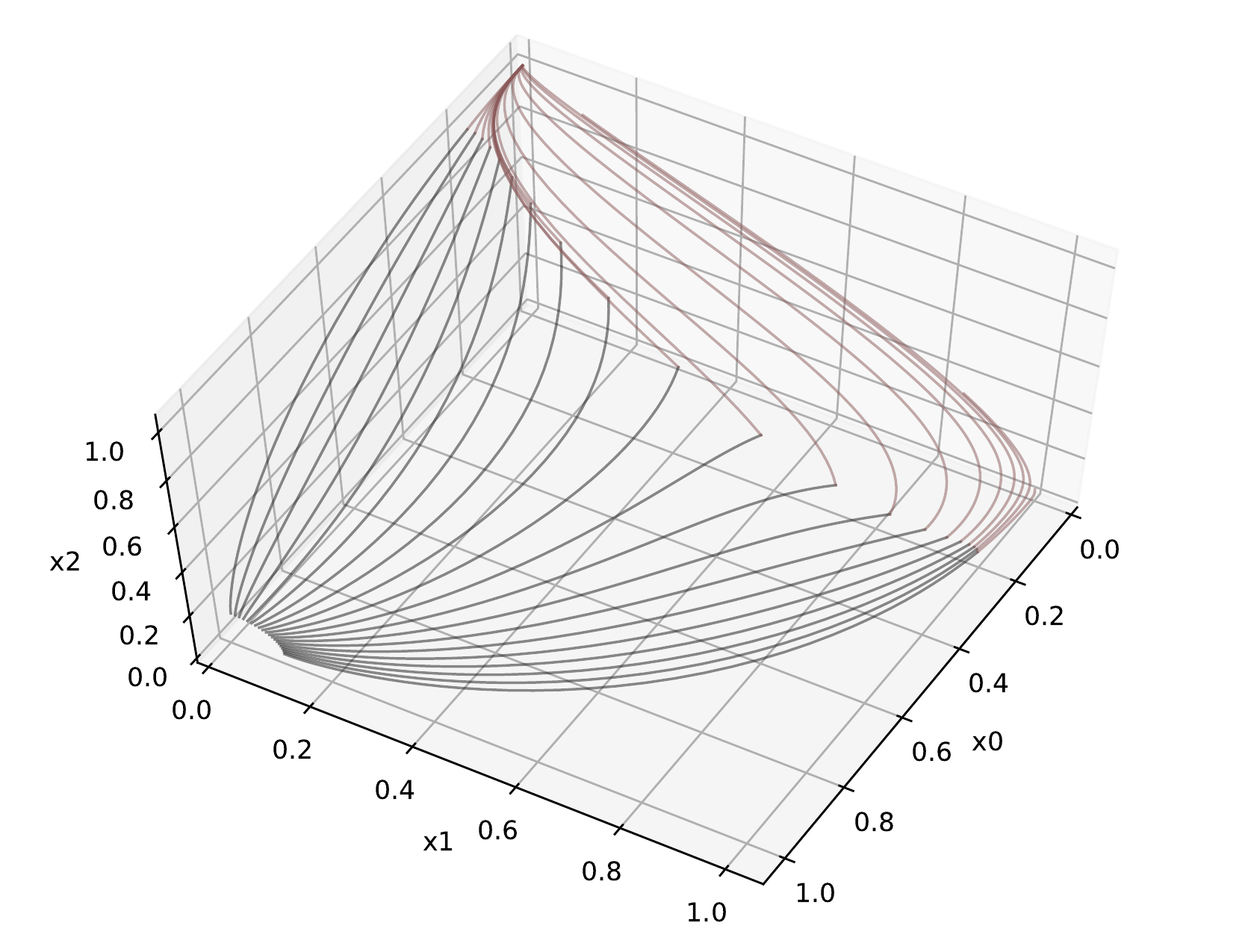}\includegraphics[width=0.33\textwidth]{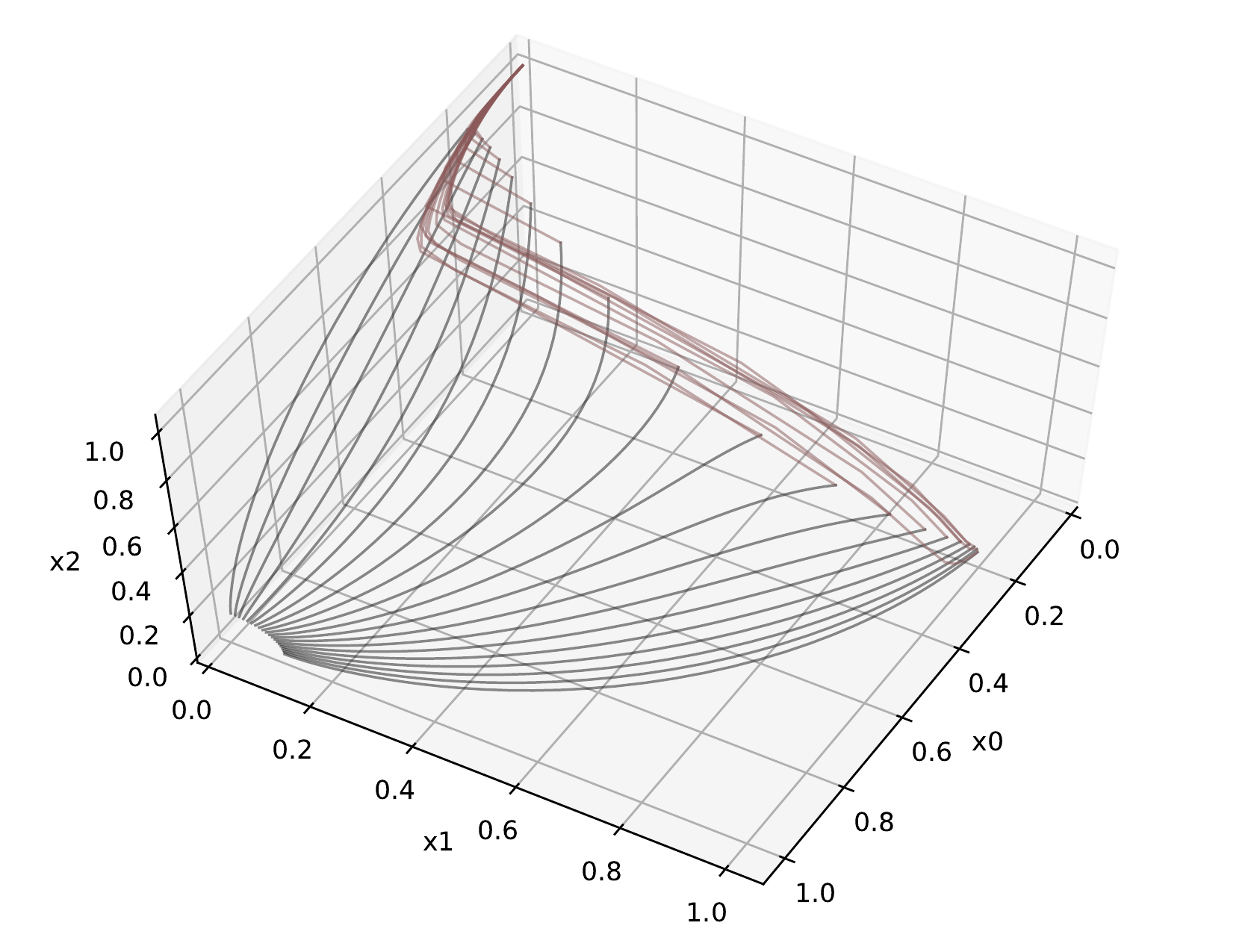}

\caption{Asymmetric greediness can be used to ``reconsider'' earlier decisions
between competing transitions. Black lines indicate where $g=[1,1,1]$,
with different initial biases during predecessor state activation.
Red lines indicate where (f.l.t.r) $g=[0,0,1]$, $g=[0,0.5,2]$, $g=[0,-1,20]$.
\label{fig:asymmetric_greediness}}
\end{figure*}

The greediness can not only be used to alter mixing behavior during
transitions, but it can also be used to make the system reconsider
the successor state it is converging to. The ratio of $\frac{g_{2}}{g_{1}}$
for two competing successor states determines where the system bifurcates.
By altering the ratio, a system that previously was set to converge
towards one state can be made to converge towards another state. This
effectively enables us to reconsider earlier decisions on which successor
state to converge to. For illustration, Fig.~\ref{fig:asymmetric_greediness}
shows three systems where during a transition, elements of $g$ are
changed asymmetrically. Depending on the absolute values, the system
can be made to ``reluctantly'' move towards the new desired successor
state (Fig.~\ref{fig:asymmetric_greediness}a), to respond gradually
depending on how certain it was before (Fig.~\ref{fig:asymmetric_greediness}b),
or to aggressively ``backtrack'' (Fig.~\ref{fig:asymmetric_greediness}c).

It should be noted, that the greediness input provides a powerful
method to alter the ``flavor'' of transitions, while keeping the
overall state graph intact (as expressed by the transition matrix
T).

\begin{figure}
\includegraphics[width=1\columnwidth]{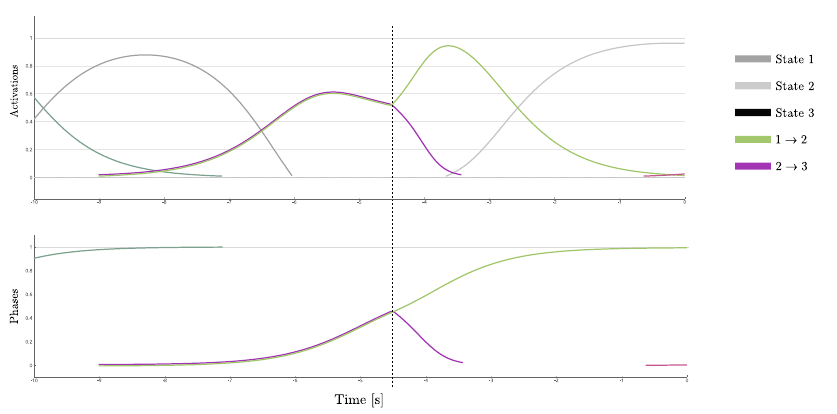}\caption{Effect of changing greediness on phases and activations. The dotted
line indicates where $g=[1,1,1]$ changes to $g=[1,5,0]$.}

\end{figure}

\section{Experiments}

In order to demonstrate the capability of the proposed system to quickly
react to perceptual input, and to generate legible motion, all while
maintaining a simple state graph abstraction of an interaction, we
chose to apply it for a handover task. In this task, a robot arm picks
up an object from a table surface, and then hands it over to a human
interaction partner standing nearby. The object can be handed over
to the left or the right hand of the human

illustrated the wealth of behaviors that we can implement with non-instant
transitions between markovian states, such as modulating decisiveness/hesitation,
reconsideration of decisions after the transition has started, and
even aborting ongoing transitions. In the context of human-robot interaction,
these behaviors enable communication by body language for negotiating
interaction alternatives and for synchronizing actions. In order to
demonstrate the feasibility of the system to generate legible motions,
and to implement negotiability of interaction alternatives, we implement
a handover task. The robot picks up an object and then has two options:
it can give the object either into the right hand or the left hand
of the human interaction partner. The human can indicate his/her preference
by extending or retracting the respective hand, giving four possible
options. If no hand is extended, then the human does not communicate
any preference. If either hand is extended, the preference is clear.
If both hands are extended, the robot interprets it as an offer to
choose; either hand is fine, and the robot should choose swiftly.
Additionally, we use the distance and orientation of the humans torso
to gauge their readiness for interaction. If a human turns away or
walks away, an ongoing reachout by the robot needs to be aborted.

The proposed system provides a consistent method to generate legible
mixtures of phase based motions such as probabilistic movement primitives
\cite{paraschos_probabilistic_2013} or dynamic movement primitives\cite{schaal_dynamic_2006}

\section{Discussion}

\subsection{Conclusions}

The paper presented a novel method to structure and execute robot
motion, which is especially suited for implementing human-robot interaction
based on body motion. The paper analyzed properties and parameters
of the proposed system and related modulations to human-interpretable
qualities such as decisiveness and hesitation, which can be used to
negotiate decisions faster and more effectively than relying on turn-based
interaction.

\section*{References}

\bibliographystyle{plain}
\bibliography{IEEEabrv,amsj,mti-engage}

\end{document}